\documentclass[lettersize,journal]{IEEEtran}
\usepackage{amsmath,amsfonts}
\usepackage{algorithmic}
\usepackage{algorithm}
\usepackage{array}
\usepackage[caption=false,font=normalsize,labelfont=sf,textfont=sf]{subfig}
\usepackage{textcomp}
\usepackage{stfloats}
\usepackage{url}
\usepackage{verbatim}
\usepackage{graphicx}
\usepackage{cite}

\usepackage{bm, graphicx}
\usepackage{multirow}
\usepackage{epstopdf}
\usepackage{etoolbox}
\usepackage{amssymb}
\usepackage{booktabs}
\begin{document}

\title{DiriNet: A network to estimate the spatial and spectral degradation functions}

\author{Ting~Hu

\thanks{T. Hu is with the Beijing Key Laboratory of Fractional Signals and Systems, School of Information and Electronics, Beijing Institute of Technology, Beijing 100081, China (3120170379@bit.edu.cn).
}}






\maketitle

\begin{abstract}
The spatial and spectral degradation functions are critical to hyper- and multi-spectral image fusion. However, few work has been payed on the estimation of the degradation functions. To learn the spatial response function and the point spread function from the image pairs to be fused, we propose a Dirichlet network, where both functions are properly constrained. Specifically, the spatial response function is constrained with positivity, while the Dirichlet distribution along with a total variation is imposed on the point spread function. To the best of our knowledge, the neural netwrok and the Dirichlet regularization are exclusively investigated, for the first time, to estimate the degradation functions. Both image degradation and fusion experiments demonstrate the effectiveness and superiority of the proposed Dirichlet network.
\end{abstract}


\section{Introduction}\label{sec:Sec1}
Limited by the tradeoff between the spatial and spectra resolution, remote-sensing spectrometers always can capature hyper-spectral images (HSI) with poor spatial resolution or multi-spectral images (MSI) with poor spectral resolution. Fortunately, images with both high-spatial and high-spectral resolution become possible, as hyper- and multi-spectral image (HMI) fusion methods emerg. Whether modeling-based \cite{2021Hyperspectral,2012Coupled} or deep learning-based \cite{2020FusionNet,2018Unsupervised} fusion techniques usually depend on the point spread function (PSF) and the spectral response function (SRF). Endless studies on the HMI fusion remain now, but few work is publised to estimate the PSF and SRF.

Generally, the observed HSI $\mathcal{X} \in \mathbb{R}^{m \times n \times B}$ and MSI $\mathcal{Y} \in \mathbb{R}^{M \times N \times b}$ are modeled as the spatial and spectral degradations of the ground-truth target $\mathcal{Z} \in \mathbb{R}^{M \times N \times B}$, respectively. Namely,
\begin{equation}\label{eq:1}
\begin{aligned}
\mathcal{X} &= D(\mathcal{Z} \ast \boldsymbol{\Phi}),\\
&\mathcal{Y} = \mathcal{Z}_{\times 3} \mathbf{R}^\mathrm{T},
\end{aligned}
\end{equation}
where $D(\cdot)$ denotes the downsampling operator of interval $r$, $r = \frac{M}{m} = \frac{N}{n}$, $\boldsymbol{\Phi} \in \mathbb{R}^{r \times r}$ is the PSF, and $\mathbf{R} \in \mathbb{R}^{B \times b}$ is the SRF. The linear relationship between the observed HSI and MSI is easily inferred as
\begin{equation}\label{eq:2}
\mathcal{X}_{\times 3} \mathbf{R}^\mathrm{T} = D(\mathcal{Y} \ast \boldsymbol{\Phi}).
\end{equation}
Obviously, the estimation of the degradation functions is a mathmatic ill-posed problem. One feasible methodology for solving this problem is to impose priori constrains on the PSF and SRF \cite{2004Recovering}. Sim$\tilde{o}$es \textit{et al}. published a subspace-based hyperspectral superresolution (HySure) \cite{2015A}, where the PSF was estimated under the co-constrain of the sum-to-one property and the spatial smoothness, while the SRF was estimated by minimizing its horizontal variation. To reduce the computational burden, the HSI and MSI are blurred certainly and the SRF and PSF are then estimated successively. Although this optimized method is often used in HMI fusion, it is revealed with worse effect when applied in the case of limited SRF overlaps \cite{2017Hyperspectral}. Finlayson's \cite{1998Recovering} method that expressed the smoothness of SRF using a finite linear combination of band-limited bases, was found to be suitable for such limited cases \cite{2017Hyperspectral}. However, it work only when PSF is konwn.

Attractived by the superior learning ability of neural networks\cite{Wei2017Transferred,Zhang2018Diverse}, we do the first attempt to design a Dirichlet network (DiriNet) to learn the PSF and SRF from pairs of HSI and MSI. In the proposed network, the spatial blurring and spectral degradation are implemented by the 2D convolution and depthwise separable convolution \cite{2017Xception}, respectively. When completing the training for DiriNet, the convolution filters are dertaimed. Then, the SRF is the dimension-squeeze result of the 2D convolution filter, and the PSF can be obtain by taking anyone piece from the depthwise convolution filter. To effectively constrain the solution space of the PSF and SRF, several regulariations are applied. Under the first discovery that PSF could be expressed by a Dirichlet distributed sequence, the stick-breaking process \cite{1994A} are used to calculate it. Moreover, the PSF is further constrained smoothingly by a total variation. As for the SRF based 2D convolution filter, it is  activated as a positive tensor via the softplus function. To optimize the DiriNet, a cost function consisting of the above variation regularization and the mean square error (MSE) between both sides of Eq.~\eqref{eq:2} is established. 

The reminder of the paper is arranged as follows. The proposed DiriNet is expatiated in Sec.~\ref{sec:Sec2} with experimental analysis being presented in Sec.~\ref{sec:Sec3}. Finally, a summary is given in Sec.~\ref{sec:Sec4}.

\begin{figure}[htbp]
	\centering
	\includegraphics[width=0.48\textwidth,trim=90 300 60 135, clip]{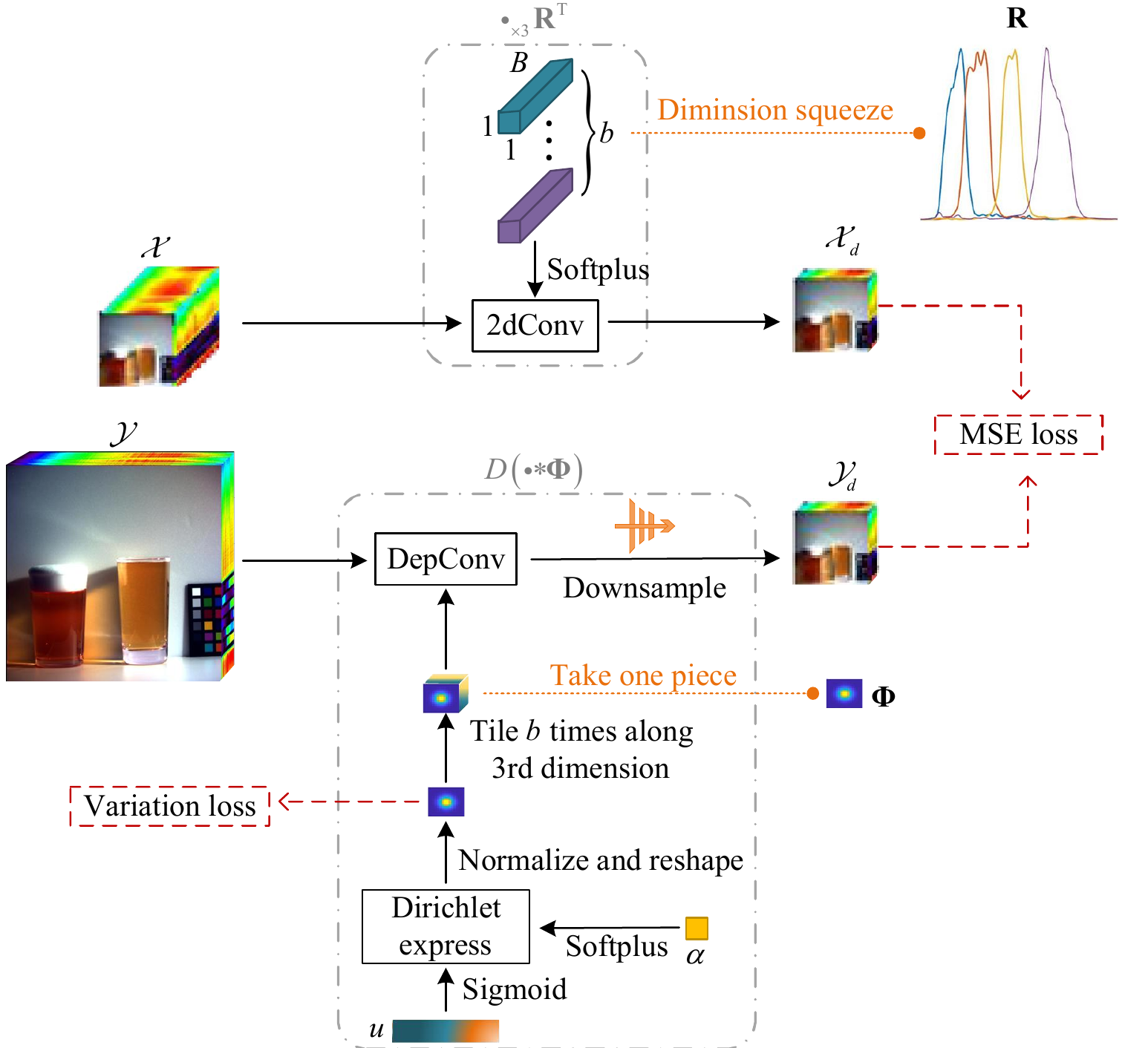}
	\caption{Flowchart of DiriNet. (The network architecture follows black arrows; orange dotted lines output the learned SRF and PSF.)}
\end{figure}\label{Fig1}

\section{Proposed DiriNet}\label{sec:Sec2}
For solving the ill-posed problem \eqref{eq:2}, DiriNet shown in Fig.~\ref{Fig1} is proposed to learn the SRF and PSF. The keys of DiriNet contains the spatial and spectral degradation process.

In the TensorFlow framework, the spectral degeneration that multiplys an image tensor with a spectrum matrix in  $ \mathbb{R}^{b \times B}$ along the spectral mode, can be easily implemented by the 2D convolution of the image and a $1 \times 1 \times B \times b$ spectrum filter \cite{2020MHF}. SRF is positive, so the spectrum filter is activated with the softplus function. Based on these, the spectral degradation is constructed by the cascade of an activation function and a 2D convolution layer, highlighted as the  dotted box above in Fig.~\ref{Fig1}.

As Eq.~\eqref{eq:1} expressed, the spatial bluring is always modeled as a band-wise convolution followed by the downsampling operation. The depthwise separable convolution proposed by Chollet \cite{2017Xception}  can just accomplish such band-wise blurring by convoluting the image with a $r \times r \times b \times 1$ filter, where the filter is an along-channel tiling of the PSF. To ensure a proper network convergence, the PSF-based filter should be constrained with some priori. The unignorable priori for PSF is that $\phi_i \in (0,1)$ and $\sum\limits_{i=1}^{r^2} \phi_i = 1$, where $\phi_i$ denotes the $i$th element of $\boldsymbol{\Phi}$. Such priori is successfully incorprated via the Dirichlet distribution in \cite{2018Unsupervised}. Similarily, the PSF is encourage to follow a Dirichlet distribution here. A random sequence $\{s_1, s_2, \cdots, s_{r^2}\}$ following the Dirichlet distribution could be produced via stick-breaking process \cite{1994A}. Specifically,
\begin{equation}\label{eq:3}
s_i = \left\{
\begin{array}{lr}
v_1  & i = 1 \\
v_i\prod_{k=1}^{i-1}(1-v_k) & \rm others
\end{array}
\right.,
\end{equation}
where $v \sim \rm Beta(1, \alpha)$ with $\alpha > 0$ being a parameter of the Beta distribution. According to the closed-form CDF of $\rm Beta(1, \alpha)$, there is
\begin{equation}\label{eq:4}
v = 1 - u^{\frac{1}{\alpha}},
\end{equation}
where $u \sim \rm Uniform(0,1)$ and $\alpha$ are learned through training the proposed DiriNet. Following the priori of $\alpha$ and $u$, they are activated with the softplus and sigmoid functions, respectively. It is hard to guarentee $u$ learned from data sets to be randomly uniform, so we impose an extra sum-to-one constrain on PSF by a normalizated operation $\phi_i = s_i / \sum\limits_{i=1}^{r^2}s_i$.

To further limit the solution space of the linear problem \eqref{eq:2}, PSF is smoothingly constrained with a total-variation loss $l_v$.
\begin{equation}\label{eq:5}
l_v = ||\bigtriangledown_x \boldsymbol{\Phi}||_1 + ||\bigtriangledown_y \boldsymbol{\Phi}||_1,
\end{equation}
where $\bigtriangledown_x$ and $\bigtriangledown_y$ express the horizontal and vertical gradient operators, respectively.

Following Eq.~\eqref{eq:1}, the spectral degradation of HSI $\mathcal{X}_d$ should be equal to the spatial degradation of MSI $\mathcal{Y}_d$. Hence, an MSE loss is constructed as
\begin{equation}\label{eq:6}
l_m = \frac{1}{mnb}||\mathcal{X}_d - \mathcal{Y}_d||_F^2.
\end{equation}
Combining the two loss using a proper regularization parameter $\lambda$, the cost function to update DiriNet is obtained
\begin{equation}\label{eq:7}
l = l_m + \lambda l_v.
\end{equation}
Once the DiriNet update is over, its weights that contains a $1 \times 1 \times B \times b$ spectrum filter, a sequence $u$, and a positive number $\alpha$, are obtained. Then, SRF is a dimension-squeeze version of the spectrum filter, and PSF could be calculated by $u$ and $\alpha$.
\section{Experimental Analysis}\label{sec:Sec3}
\begin{figure*}[htbp]
\centering
\subfloat[]{\includegraphics[width=0.23\textwidth, trim=2 10 25 15, clip]{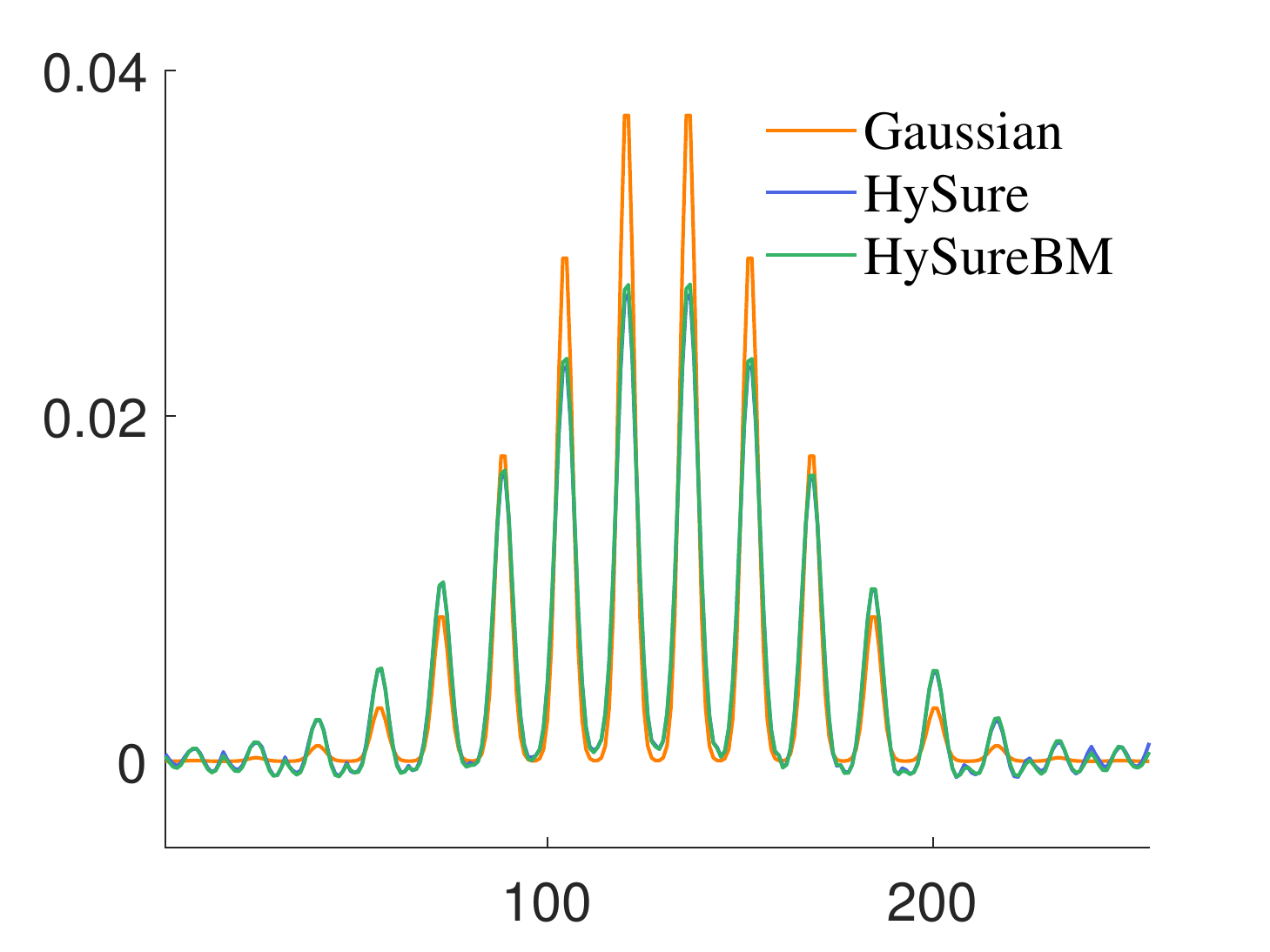}
\label{Fig2sub1}}
\hfil
\subfloat[]{\includegraphics[width=0.23\textwidth, trim=2 10 25 15, clip]{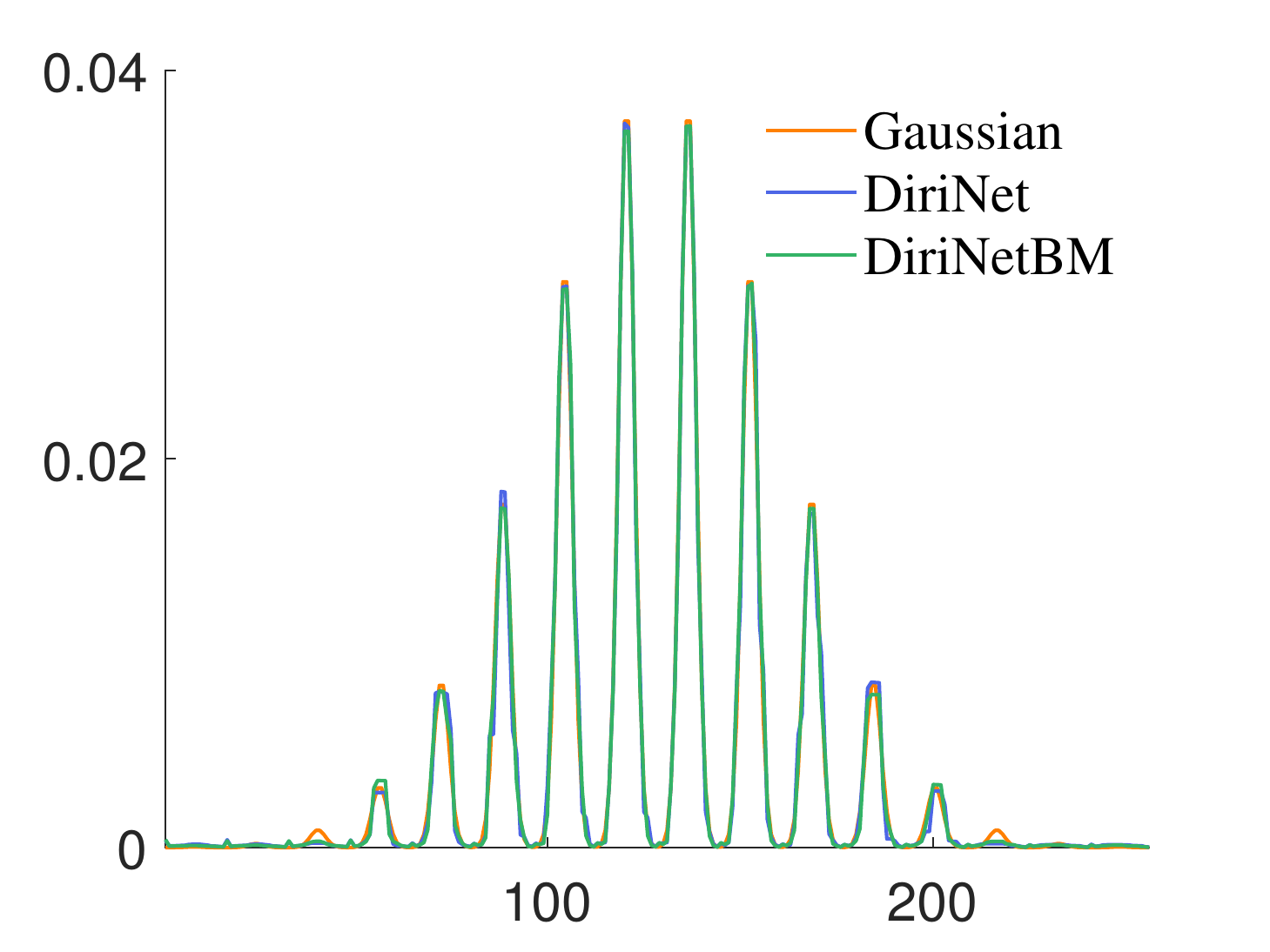}
\label{Fig2sub2}}
\hfil
\subfloat[]{\includegraphics[width=0.23\textwidth, trim=2 10 25 15, clip]{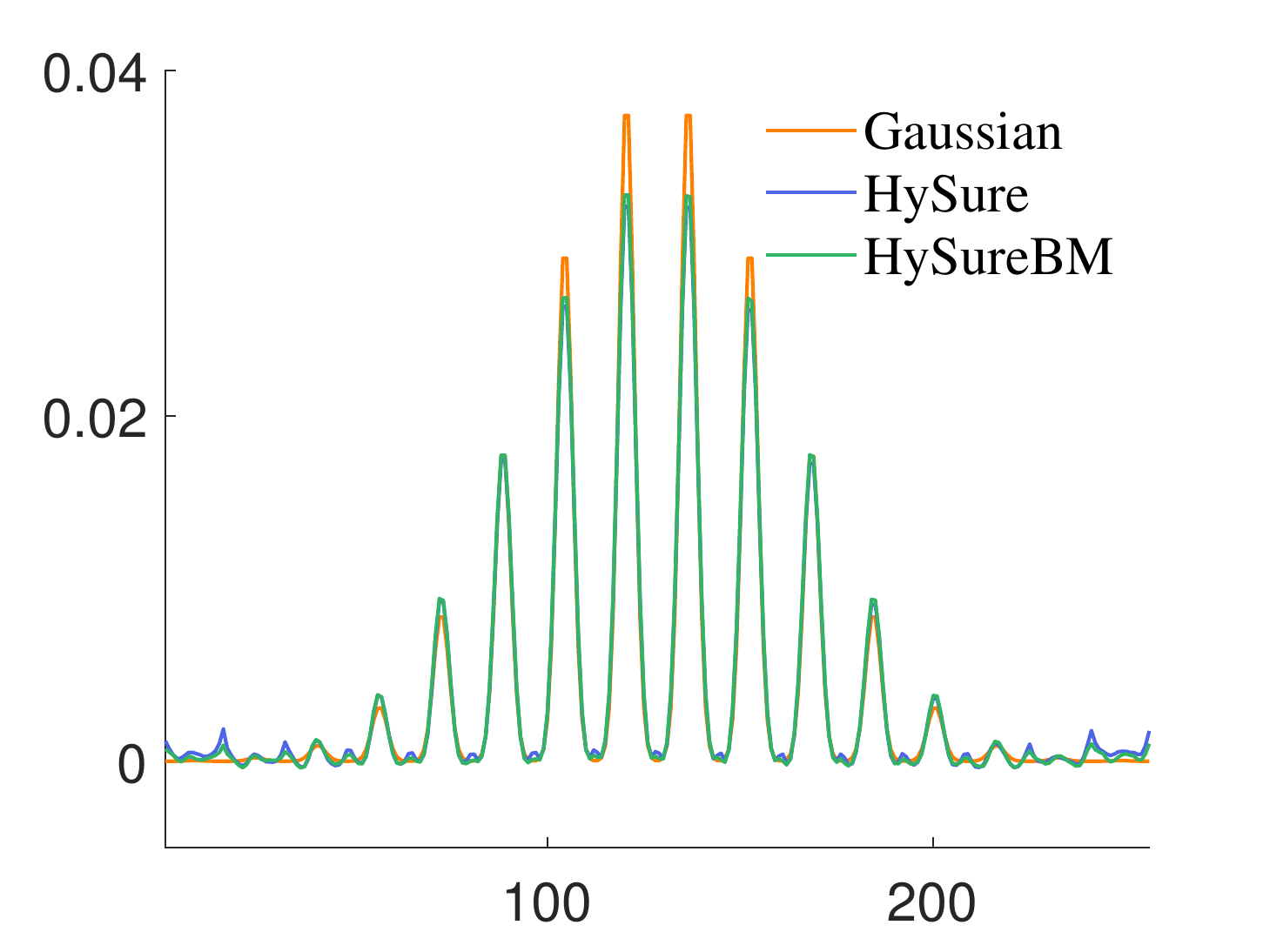}
\label{Fig2sub3}}
\hfil
\subfloat[]{\includegraphics[width=0.23\textwidth, trim=2 10 25 15, clip]{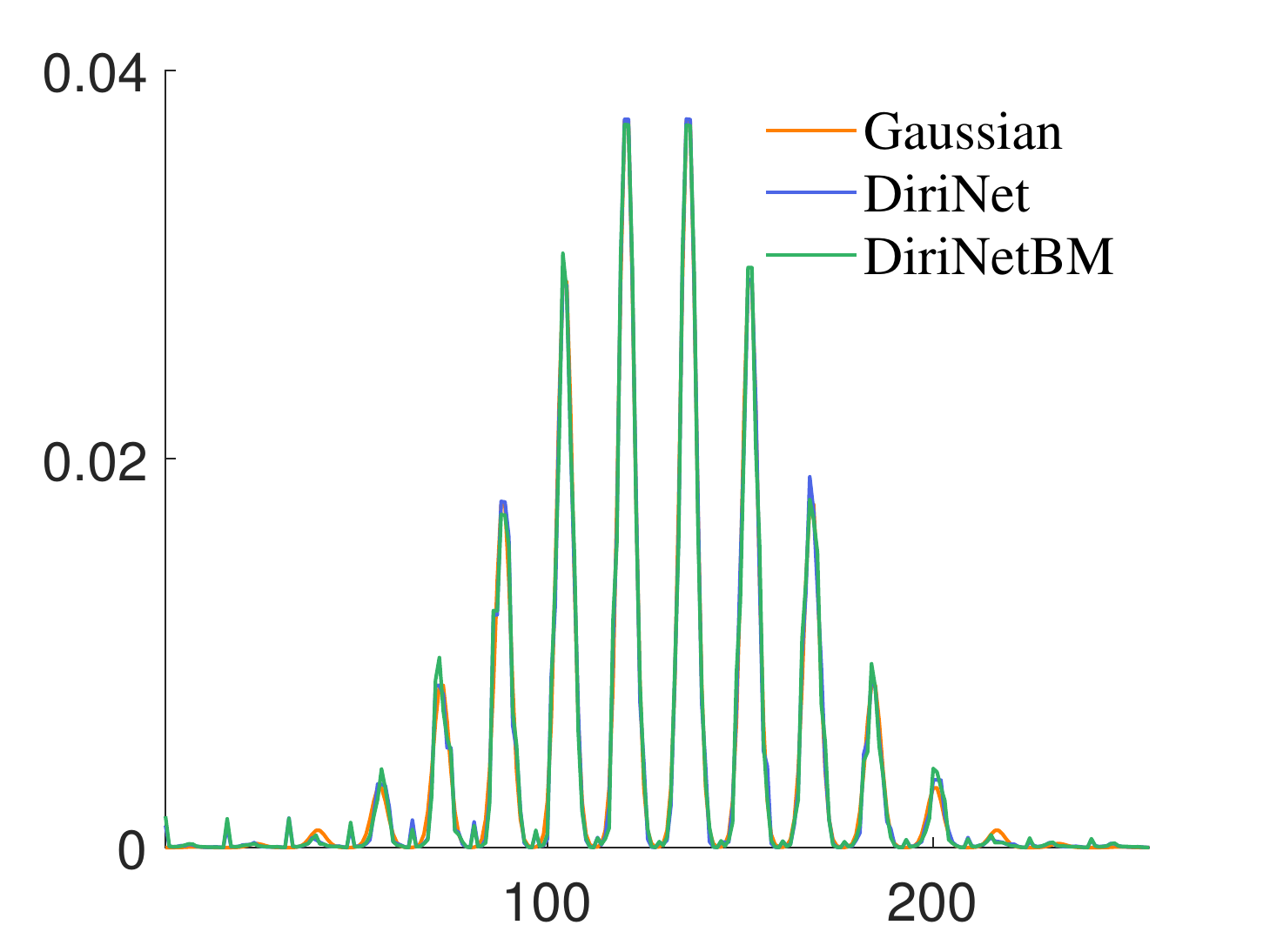}
\label{Fig2sub4}}

\subfloat[]{\includegraphics[width=0.23\textwidth, trim=2 10 25 15, clip]{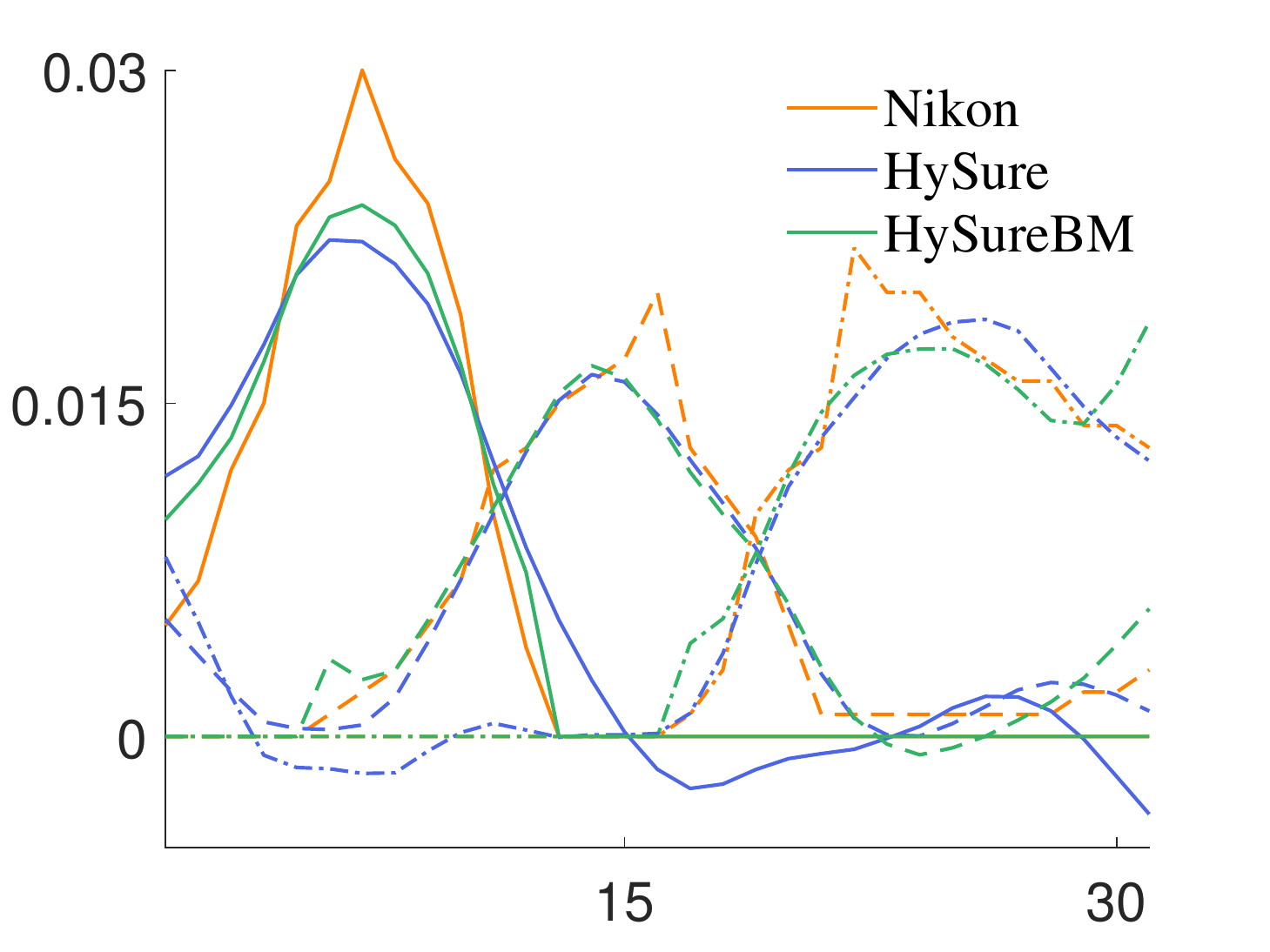}
  \label{Fig2sub5}}
\hfil  
\subfloat[]{\includegraphics[width=0.23\textwidth, trim=2 10 25 15, clip]{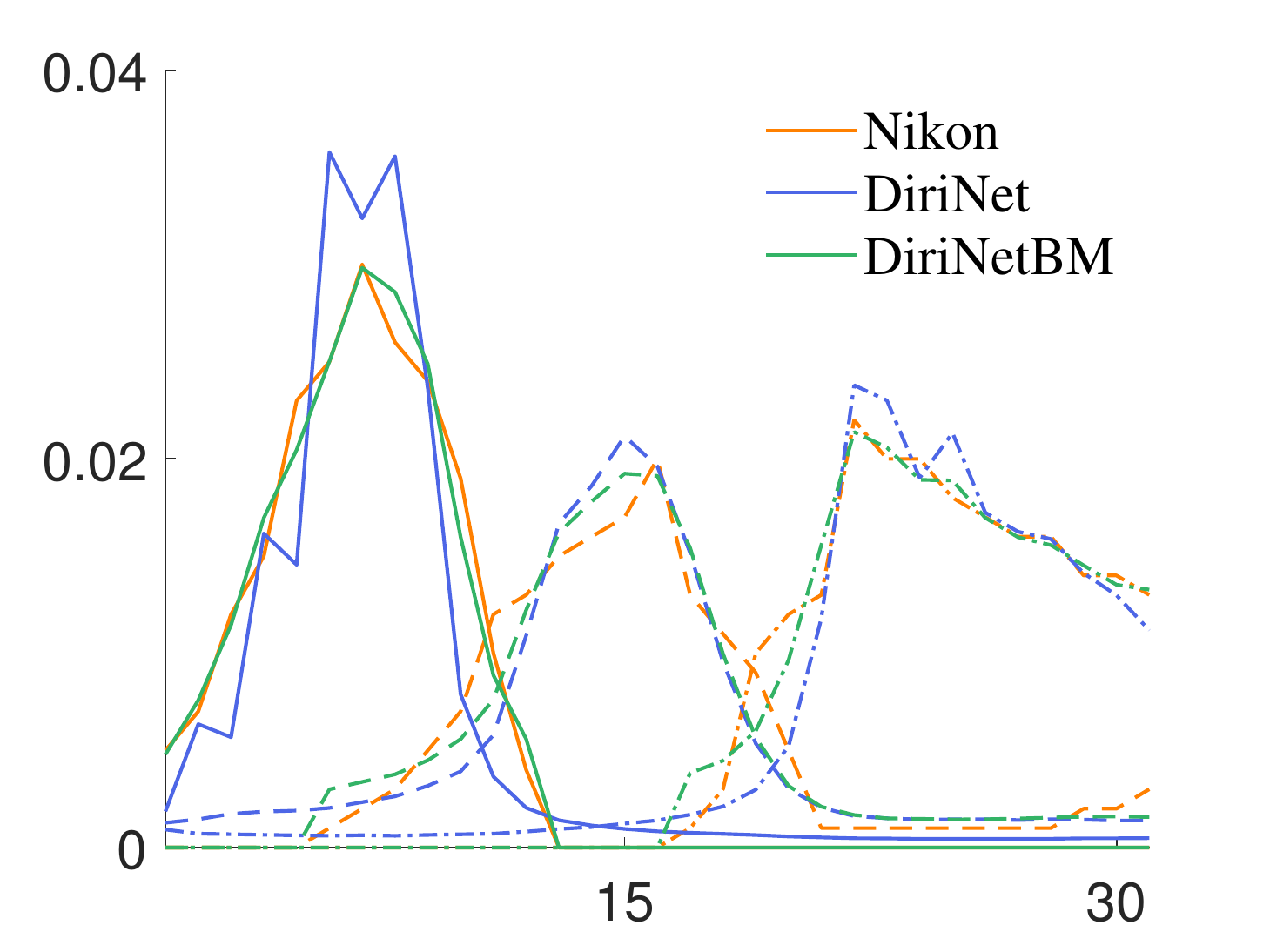}
\label{Fig2sub6}}
\hfil
\subfloat[]{\includegraphics[width=0.23\textwidth, trim=2 10 25 15, clip]{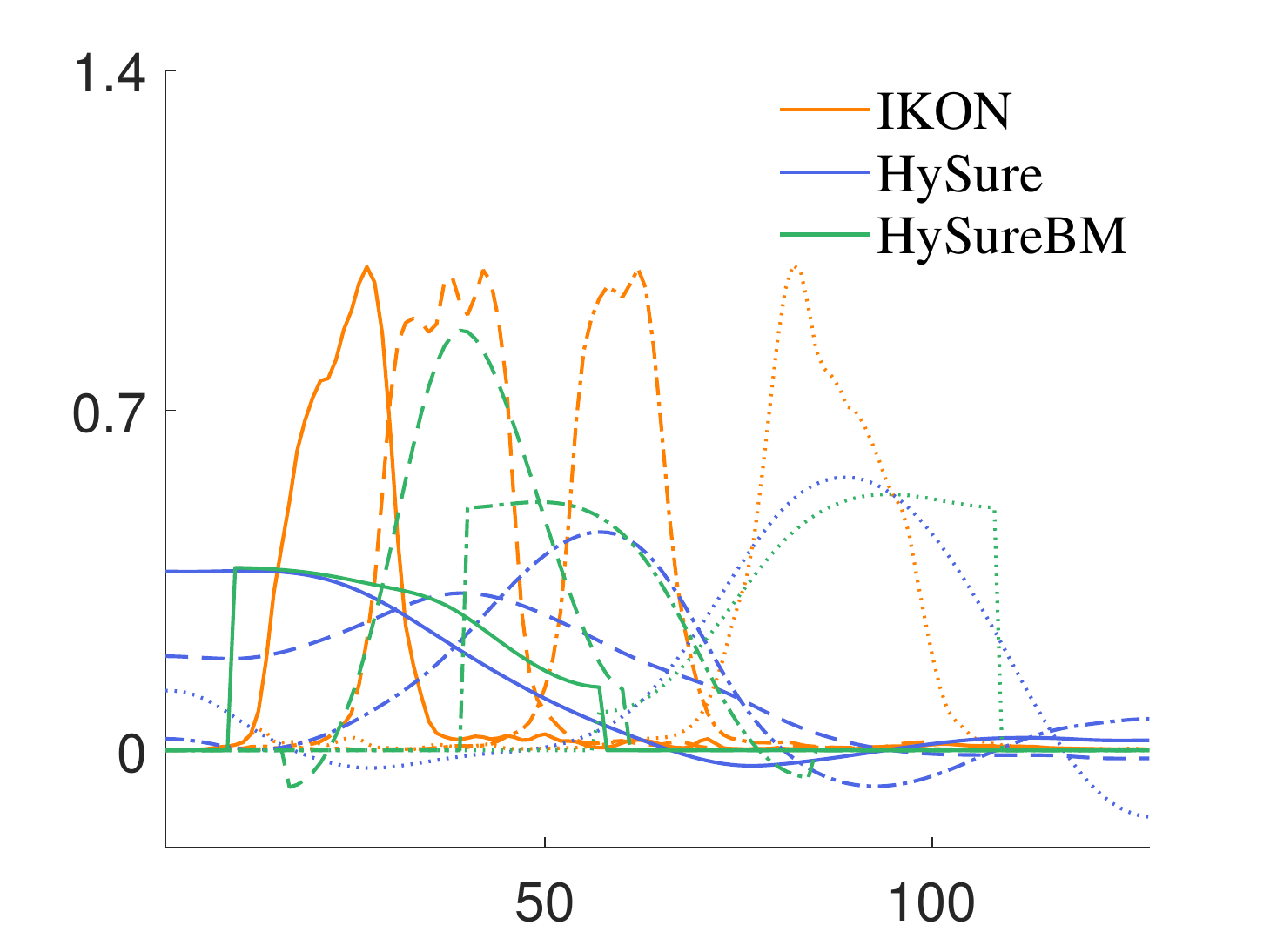}
\label{Fig2sub7}}
\hfil  
\subfloat[]{\includegraphics[width=0.23\textwidth, trim=2 10 25 15, clip]{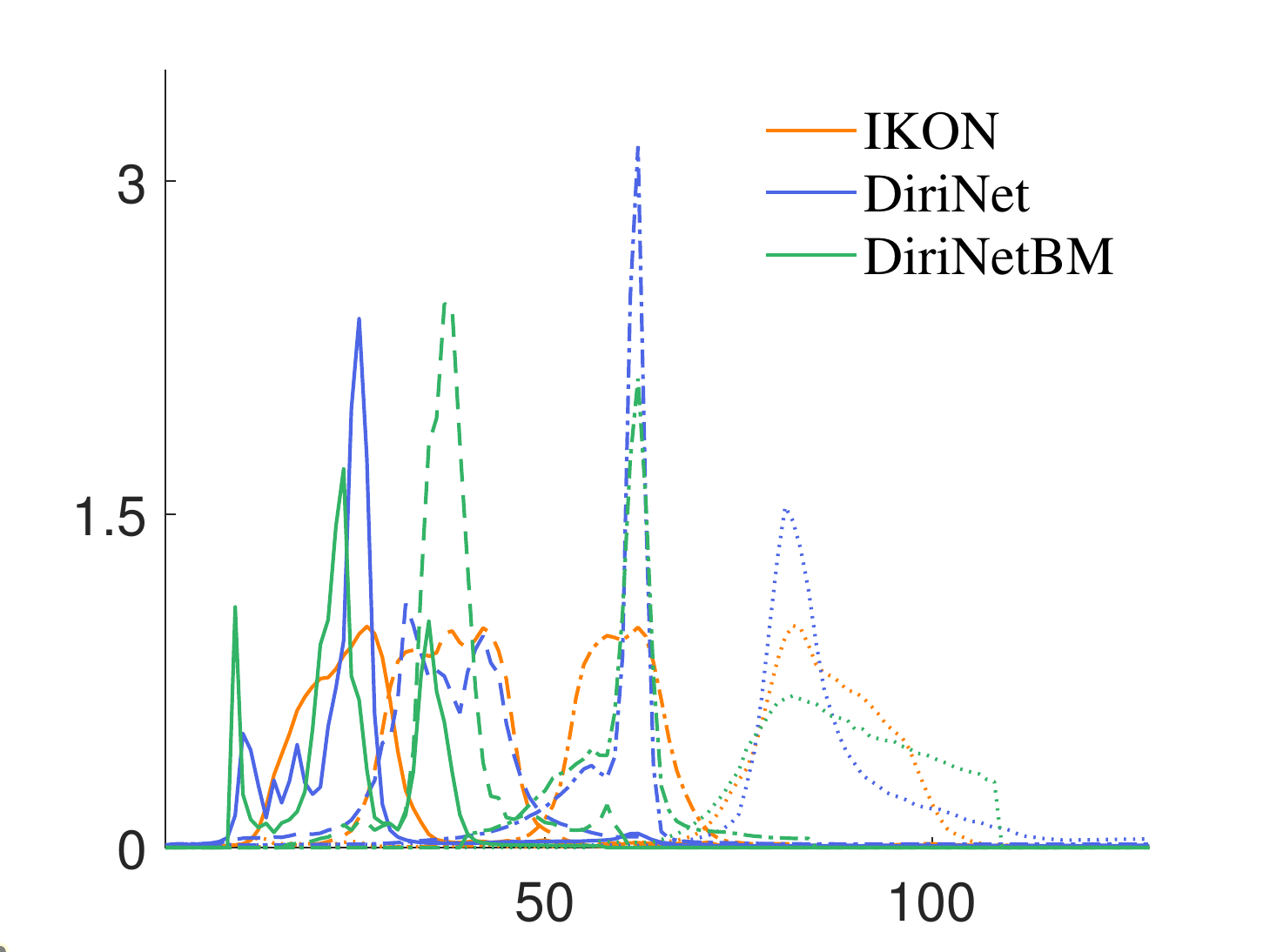}
\label{Fig2sub8}}

\caption{Illustration of PSF and SRF. (a)-(b) and (c)-(d) plot the PSF of Cave1 and Chikusei1, respectively, where $16 \times 16$ PSF is reshaped as a sequence of length $256$. (e)-(f) and (g)-(h) plot the SRF of Cave1 and Chikusei1, respectively, where curves of different types depict SRF in different bands.}
\label{Fig2}
\end{figure*}

\begin{table*}[h]
	\caption{Metrics between Cave1-2's degradations obtained by estimated and original responses\label{tab:Tab1}}
	\centering
		\begin{tabular}{|c|c|cccccc|cccccc|}
			\hline
			\multirow{2}{*}{Response} & \multirow{2}{*}{Method} & \multicolumn{6}{c|}{$16 \times 16$ Gaussian kernel} & \multicolumn{6}{c|}{$32 \times 32$ Average kernel}  \\ \cline{3-14}
			&                         & PSNR    & SSIM   & ERGAS  & Q$2^n$ & SAM    & SID    & PSNR    & SSIM   & ERGAS  & Q$2^n$  & SAM    & SID    \\ \hline
			\multirow{4}{*}{SRF}      & HySure                  & 61.25 & 0.9987 & 0.3226 & 0.9526 & 1.4900  & 1.8092 & 54.18 & 0.9950 & 0.3880 & 0.8813 & 4.6430 & 8.3311 \\
			& HySureBM                & 64.14 & 0.9994 & 0.2594 & 0.9796 & 1.3641 & 0.7326 & 57.95 & 0.9976 & 0.3054 & 0.9252 & 2.3564 & 3.2015 \\
			& DiriNet                 & 65.51 & 0.9992 & 0.2411 & 0.9799 & 2.4959 & 1.0165 & 65.56 & 0.9992 & 0.1195 & 0.9793 & 2.3803 & 0.9677 \\
			& DiriNetBM               & 83.46 & 1.0000 & 0.0472 & 0.9989 & 0.2727 & 0.0255 & 81.66 & 1.0000 & 0.0249 & 0.9985 & 0.2780 & 0.0278 \\ \hline
			\multirow{4}{*}{PSF}      & HySure                  & 53.09 & 0.9995 & 0.1640 & 0.9997 & 0.3992 & 1.6409 & 50.35 & 0.9992 & 0.1117 & 0.9994 & 0.4235 & 2.5793 \\
			& HySureBM                & 54.48 & 0.9996 & 0.1408 & 0.9998 & 0.3543 & 1.2371 & 55.36 & 0.9997 & 0.0628 & 0.9998 & 0.2398 & 0.8205 \\
			& DiriNet                 & 72.62 & 1.0000 & 0.0180 & 1.0000 & 0.0801 & 0.0248 & 71.38 & 1.0000 & 0.0105 & 1.0000  & 0.0487 & 0.0263 \\
			& DiriNetBM               & 75.64 & 1.0000 & 0.0126 & 1.0000 & 0.0533 & 0.0131 & 89.73 & 1.0000 & 0.0012 & 1.0000 & 0.0059 & 0.0006 \\ \hline
		\end{tabular}
\end{table*}

\begin{table*}[h]
	\caption{Metrics between Chikusei1-2's degradations obtained by estimated and original responses\label{tab:Tab2}}
	\centering
		\begin{tabular}{|c|c|cccccc|cccccc|}
			\hline
			\multirow{2}{*}{Response} & \multirow{2}{*}{Method} & \multicolumn{6}{c|}{$16 \times 16$ Gaussian kernel} & \multicolumn{6}{c|}{$32 \times 32$ Average kernel}   \\ \cline{3-14}
			&                         & PSNR    & SSIM   & ERGAS  & Q$2^n$  & SAM    & SID     & PSNR    & SSIM   & ERGAS  & Q$2^n$ & SAM    & SID     \\ \hline
			\multirow{4}{*}{}         & HySure                  & 40.99 & 0.9958 & 0.3873 & 0.9918 & 1.1325 & 9.7652  & 36.29 & 0.9864 & 0.3176 & 0.9843 & 2.2391 & 38.9938 \\
			& HySureBM                & 40.40 & 0.9937 & 0.4306 & 0.9879 & 1.2977 & 17.9865 & 38.36 & 0.9856 & 0.3418 & 0.9738 & 2.3097 & 43.9317 \\
			& DiriNet                 & 55.28 & 0.9998 & 0.0676 & 0.9994 & 0.1784 & 0.4724  & 51.54 & 0.9994 & 0.0572 & 0.9983 & 0.2783 & 1.1288  \\
			& DiriNetBM               & 53.85 & 0.9993 & 0.1102 & 0.9980 & 0.2979 & 1.3603  & 52.47 & 0.9992 & 0.0592 & 0.9980 & 0.3251 & 1.4748  \\ \hline
			\multirow{4}{*}{PSF}      & HySure                  & 74.19 & 0.9999 & 0.1129 & 0.9995 & 0.2749 & 0.1635  & 78.42 & 1.0000 & 0.0357 & 0.9997 & 0.1578 & 0.0377  \\
			& HySureBM                & 77.33 & 1.0000 & 0.0791 & 0.9998 & 0.1849 & 0.0650  & 79.44 & 1.0000 & 0.0318 & 0.9998 & 0.1391 & 0.0313  \\
			& DiriNet                 & 86.62 & 1.0000 & 0.0275 & 1.0000 & 0.0748 & 0.0335  & 89.61 & 1.0000 & 0.0105 & 1.0000 & 0.0398 & 0.0027  \\
			& DiriNetBM               & 86.09 & 1.0000 & 0.0292 & 1.0000 & 0.0787 & 0.0370  & 89.22 & 1.0000 & 0.0109 & 1.0000 & 0.0404 & 0.0028  \\ \hline
		\end{tabular}
\end{table*}

\begin{table*}[htbp]
	\caption{Fusion metrics on Cave1-2 and Chidusei1-2 for different estimation methods\label{tab:Tab3}}
	\centering
	\begin{tabular}{|c|c|cccccc|cccccc|}
			\hline
			\multirow{2}{*}{Image}    & \multirow{2}{*}{Responses} & \multicolumn{6}{c|}{$16 \times 16$ Gaussian kernel}     & \multicolumn{6}{c|}{$32 \times 32$ Average kernel}      \\ \cline{3-14}
			&                            & PSNR    & SSIM   & ERGAS  & Q$2^n$ & SAM    & SID     & PSNR    & SSIM   & ERGAS  & Q$2^n$    & SAM     & SID      \\ \hline
			\multirow{5}{*}{Cave}     & HySure                     & 41.64 & 0.9818 & 0.8312 & 0.9303 & 6.5101 & 36.2442 & 39.67 & 0.9724 & 1.0027 & 0.8882 & 11.4057 & 159.9813 \\
			& HySureBM                   & 41.74 & 0.9816 & 0.7945 & 0.9288 & 6.5166 & 35.8018 & 41.93 & 0.9800 & 0.3771 & 0.9234 & 7.3714  & 42.1367  \\
			& DiriNet                    & 41.81 & 0.9815 & 0.8426 & 0.9269 & 6.6587 & 36.6442 & 42.35 & 0.9796 & 0.3744 & 0.9179 & 7.6670  & 42.3063  \\
			& DiriNetBM                  & 42.22 & 0.9829 & 0.8210 & 0.9349 & 6.2072 & 34.5704 & 42.68 & 0.9813 & 0.3557 & 0.9270 & 7.1715  & 39.1826  \\
			& Nikon                      & 42.08 & 0.9827 & 0.8243 & 0.9363 & 6.2972 & 34.5135 & 42.62 & 0.9812 & 0.3590 & 0.9284 & 7.0334  & 38.8715  \\ \hline
			\multirow{5}{*}{Chikusei} & HySure                     & 62.08 & 0.9987 & 0.8036 & 0.8861 & 6.4752 & 2.5291  & 62.85 & 0.9987 & 0.3830 & 0.8996 & 6.5281  & 2.6552   \\
			& HySureBM                   & 62.41 & 0.9986 & 1.0912 & 0.8880 & 6.5270 & 2.7833  & 62.66 & 0.9987 & 0.3854 & 0.8967 & 6.5203  & 2.6694   \\
			& DiriNet                    & 63.41 & 0.9988 & 0.7654 & 0.8861 & 6.4942 & 2.3654  & 63.67 & 0.9988 & 0.3880 & 0.8999 & 6.3565  & 2.3024   \\
			& DiriNetBM                  & 63.66 & 0.9988 & 0.7568 & 0.8872 & 6.4631 & 2.2461  & 63.71 & 0.9988 & 0.3892 & 0.9033 & 6.3760  & 2.3211   \\
			& IKONS                    & 64.01 & 0.9988 & 0.7581 & 0.8866 & 6.4494 & 2.4408  & 63.97 & 0.9989 & 0.3807 & 0.9016 & 6.3638  & 2.2592
			\\ \hline  
		\end{tabular}
\end{table*}

To assess the influence of different SRF and PSF on estimating the degradation functions, four data sets are constructed. The CAVE data set that contains 32  images of size $512 \times 512 \times 31$, is spectrally degenerated using the SRF of Nikon D700 and spatially degenerated using a zero-mean Gaussian PSF of stardand devariation $2$ and size $16 \times 16$ and an average PSF of size $32 \times 32$. Then, two CAVE data sets (named as Cave1 and Cave2 in turn) are obtained by different PSFs. In addition, the Chikusei image comprosing $128$-band spectrum and covering $2517 \times 2335$-pixel scene is cut to 25 subimages of size $512 \times 512 \times 128$. Similarily, an IKONOS-like SRF and the above PSF are imposed on the subimages to produce Chikusei1 and Chikusei2. Obviously, Cave1-2 are the reflections of many SRF overlaps; Chikusei1-2 are the illustrations of limited SRF overlaps. 20 and 16 images are randomly slected from Cave1-2 and Chikusei1-2 to form four training data sets, respectively, and the remains are used for testing. Response estimations are executed on the training data sets, while performance assessments are accomplished using the testing data sets. 
 
Within 500 iterations, the proposed DiriNet is optimized via Adam method. When training, the initial learning rates that are set as $10^{-1}$ and $10^{-2}$ for Cave1-2 and Chikusei1-2, respectively, are damped exponentially with a decay step 250 and a dceay rate 0.99. For all data sets, $\lambda = 10^{-7}$. To reduce the convergence difficulty, the SRF is pre-trained 1000 times with deleting the PSF blurring.

For a comparative study, the state-of-the-art HySure method \cite{2015A} is considered. Each SRF curve is band-limited, so a manual band matching is applied in HySure. To test the effect of the band matching, HySure and DiriNet with and without band matching are compared in this paper. For convenience, the two methods with band matching are named as HySureBM and DiriNetBM, respectrively. The names are staied in cases of no band matching. To be fair, HySure and HySureBM are traversed the training data sets to generate seirs of responses for an average result. 

In Fig.2, the estimated PSF and SRF are plotted to visually present the estimation effect. Obviously, wether with or without band matching, the proposed networks are much preferable to the considered methods. More specifically, the estimated results of HySureBM and HySure are comparatively smoothing, which could be caused by the co-smoothness regularization on the PSF and SRF. Furthermore, some negative values appear in the PSF and SRF obtained by HySureBM and HySure, because no threshold constraint is employed. 

In Tables~\ref{tab:Tab1} and~\ref{tab:Tab2}, the peak signal-to-noise ratio (PSNR), structural similarity (SSIM), erreur relative globale adimensionnelle desynth\`ese (ERGAS), Q$2^n$, spectral angle mapper (SAM), and spectral information divergence (SID) are listed for a quantitative assessment. Higher PSNR, SSIM, and Q$2^n$ as well as lower ERGAS, SAM, and SID mean better performance. Much better metrics are obtained by DiriNet than HySure, espically for data sets with limited SRF overlaps (Chikusei1-2). Such superior performance could be the result of the good learning ability, moderate smoothness regularization, and positive constraint employed in DiriNet. The comparison between the experiments under the Gaussian and average PSF show that both HySure and HySureBM get a great performance reduction with the blurring degree. The performance degradation might be a consequence of the bluring operation before the response estimation. In addition, band matching indicats distinct benefit for the response estimation in Cave1-2 but slight damage in Chikusei1-2, probably because the tailing of IKONS-like response is nonzero. Actually, the tailings of SRFs are usually non zero; thus, band matching is a dispensable option. Compresensively, Dirichlet is an excellent approach to estimate the degradation functions, espically for data sets with limited SRF overlaps. 

To study the application effect of all considered methods, the estimated and original responses are applied in an outstanding HMI fusion method which is rooted on the coupled nonnegative matrix factorization (CNMF) \cite{2012Coupled}. The metrics between the fusions and the original HSI are calculated in Table~\ref{tab:Tab3}. As listed in Table~\ref{tab:Tab3}, the fusion effect under the DiriNet estimations is comparable to that under the original responses. Hence, the proposed DiriNet is proved to be practical.

\section{Conclusion}\label{sec:Sec4}
A Dirichlet estimation network was proposed to accurately provid spatial and spectral degradation functions for hyper-spectral and multi-spectral image fusion. Pure data-driven learning was difficult to work around this ill-posed estimation problem, so some effective constrains were applied. The PSF was constrained with Dirichlet distribution and spatial smoothness, while the SRF was regularized with non-negative property. Experiments revealed that the proposed network could not only better learn the degradation functions but also effectively serve the fusion technique.


\bibliographystyle{IEEEbib}
\bibliography{references}

\end{document}